\documentclass[letterpaper, 10 pt, conference]{ieeeconf}  % Comment this line out if you need a4paper

\IEEEoverridecommandlockouts                              % This command is only needed if 
\usepackage{scrextend}
\usepackage[hidelinks]{hyperref}
\usepackage{graphicx}   % for pdf, bitmapped graphics files
\usepackage{epsfig}    % for postscript graphics files
\usepackage{amsmath}    % assumes amsmath package installed
\usepackage{amssymb}    % assumes amsmath package installed
\usepackage{mathtools}  % assumes amsmath package installed
\usepackage{float}
\usepackage{booktabs,tabularx}
\usepackage[caption=false]{subfig}
\usepackage{cite}
\usepackage{accents}
\usepackage{blindtext}
\usepackage{xcolor}
\usepackage{multicol}       
\usepackage{booktabs,tabularx}
\usepackage{multirow}
\usepackage{accents}

\usepackage{amsthm}

\newtheorem{remark}{Remark}

\definecolor{ncs}{rgb}{0.18, 0.32, 0.62}

\title{\LARGE \bf
An Efficient Real-Time NMPC for Quadrotor \\ Position Control under Communication Time-Delay
}

\author{B\'arbara Barros Carlos$^{1}$, Tommaso Sartor$^{2}$, Andrea Zanelli$^{3}$, Gianluca Frison$^{3}$,\\
Wolfram Burgard$^{4}$, Moritz Diehl$^{3}$ and Giuseppe Oriolo$^{1}$% <-this % stops a space
\thanks{$^{1}$ B. B. Carlos and G. Oriolo are with Sapienza Universit\`a di Roma, Department of Information, Automation \& Management Engineering (DIAG), Italy. {\tt\footnotesize \{barros, oriolo\}@diag.uniroma1.it}}%
\thanks{$^{2}$ T. Sartor is with KU Leuven, Department of Mechanical Engineering (MECO Group), Belgium. {\tt\footnotesize tommaso.sartor@kuleuven.be}}%
\thanks{$^{3}$ A. Zanelli, G. Frison and M. Diehl are with the University of Freiburg, Department of Microsystems Engineering (IMTEK), Germany. {\tt\footnotesize \{andrea.zanelli, gianluca.frison, moritz.diehl\}@imtek.uni-freiburg.de}}%
\thanks{$^{4}$ W. Burgard is with the University of Freiburg, Department of Computer Science (IIF), Germany. {\tt\footnotesize burgard@informatik.uni-freiburg.de}}}

\begin{document}

\maketitle
\thispagestyle{empty}
\pagestyle{empty}

%===============================================================================
\begin{abstract}
The advances in computer processor technology have enabled the application of nonlinear model predictive control (NMPC) to agile systems, such as quadrotors. These systems are characterized by their underactuation, nonlinearities, bounded inputs, and time-delays. Classical control solutions fall short in overcoming these difficulties and fully exploiting the capabilities offered by such platforms. This paper presents the design and implementation of an efficient position controller for quadrotors based on real-time NMPC with time-delay compensation and bounds enforcement on the actuators. To deal with the limited computational resources onboard, an offboard control architecture is proposed. It is implemented using the high-performance software package \texttt{acados}, which solves optimal control problems and implements a real-time iteration (RTI) variant of a sequential quadratic programming (SQP) scheme with Gauss-Newton Hessian approximation. The quadratic subproblems (QP) in the SQP scheme are solved with \texttt{HPIPM}, an interior-point method solver, built on top of the linear algebra library \texttt{BLASFEO}, finely tuned for multiple CPU architectures. Solution times are further reduced by reformulating the QPs using the efficient partial condensing algorithm implemented in \texttt{HPIPM}. We demonstrate the capabilities of our architecture using the Crazyflie 2.1 nano-quadrotor.  
\end{abstract}

%===============================================================================
\section{INTRODUCTION}
\label{sec:intro}
Nonlinear model predictive control (NMPC) is an optimization-based control method that relies on the online solution of nonlinear nonconvex programs to compute an optimal feedback policy. In particular, domains characterized by sufficiently long sampling times, such as the chemical processing and oil refineries, have seen successful industrial applications of NMPC since the 80s. As the computation times associated with the solution of the underlying optimization problems can be rather long, the employment of NMPC has only recently been extended to applications where shorter sampling times are required. 

\begin{figure}[h]
    \centering
    \includegraphics[width=0.96\linewidth]{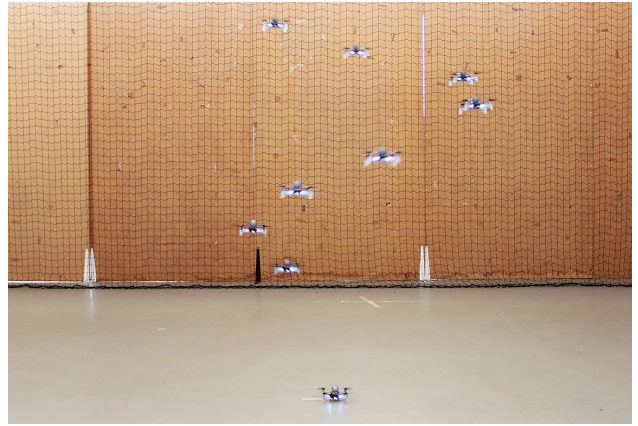}
    \caption{A composite image showing our real-time NMPC with time-delay compensation being used on the Crazyflie during the tracking of a helical trajectory. Given the current measurement, the state is predicted over the delay time interval using an integrator and then passed to the NMPC, which takes into account the input bounds. The efficiency of the proposed architecture leans on high-performance software implementations that span: real-time iteration scheme to address the NMPC problem, Hessian condensing algorithm suited for partial condensing, structure-exploiting QP solver, and hardware-tailored linear algebra library. A video of the experiments is available at {\footnotesize\color{ncs}{\url{https://youtu.be/xZLVQ7BdUHA}}}.}\label{fig:crazyflie}
\end{figure}

One important precondition for its employment in the context of agile systems is the availability of approximate schemes that enable real-time feasibility. Particularly successful schemes trade control performance for speed and typically hinge on implementations of: Newton-type method \cite{deng2020, barroscarlos2020}, augmented Lagrangian method \cite{englert2020, sathya2020}, and continuation/generalized minimal residual (C/GMRES) method \cite{guo2020, katayama2020}. Solution times can be further reduced by building efficient condensing approaches and structure-exploiting solvers. 

Quadrotors are well-known examples of agile systems. They are characterized by their underactuation, nonlinearities, bounded inputs, and, in some cases, communication time-delays. The development of their maneuvering capability poses some challenges that cover dynamics modeling, state estimation, trajectory generation, and control. The latter, in particular, must be able to exploit the system nonlinear dynamics to generate complex motions. However, the presence of communication time-delay is known to highly degrade the control performance of the system in question. This issue has been extensively studied in the literature over the years, but given the growing number of applications requiring collective interaction, most efforts are in the field of multi-agent control \cite{kartal2020, alvarez2019, huang2019}. Although these strategies are encouraging, the resulting feedback control laws are not optimal.

This paper proposes an efficient position control architecture based on real-time NMPC with time-delay compensation for quadrotors. The architecture consists of predicting the state over the delay time interval, granting a nominally delay-free model in the NMPC formulation, which takes into account the input bounds. The NMPC algorithm is built upon the real-time iteration (RTI) scheme with Gauss-Newton Hessian approximation, and it is implemented using the high-performance software package \texttt{acados} \cite{verschueren2019}. We use the automatic differentiation and modeling framework \texttt{CasADi} \cite{andersson2019} to provide sensitivity calculations. The quadratic program (QP) solver is a Riccati-based interior-point method, \texttt{HPIPM} \cite{frison2020}, which relies on the hardware-tailored linear algebra library \texttt{BLASFEO} \cite{frison2018}. We show that by using a Hessian condensing algorithm particularly well-suited for a structure-exploiting QP solver, a considerable speed-up can be achieved with respect to a state-of-the-art NMPC solver. To experimentally validate the architecture, we use the Crazyflie 2.1 nano-quadrotor, an open-source off-the-shelf platform for research and education in robotics (see Fig. \ref{fig:crazyflie} for an illustration). The main contributions of our work are: 
\begin{itemize}
	\item Provide an experimental validation of our architecture.
	\item Grant an efficient software implementation that needs to cope with resource-constrained hardware. 
\end{itemize}

This paper is outlined as follows. Section \ref{sec:nmpc_position_control} provides the details of the proposed architecture. Section \ref{sec:simulations} shows the comparative results of the simulations. Sections \ref{sec:experiments} and \ref{sec:discussion} present the experimental closed-loop results and the discussion of our strategy, respectively. In Section \ref{sec:conclusion} the conclusions are summarized.
%===============================================================================
\section{NMPC FOR POSITION CONTROL WITH TIME-DELAY COMPENSATION}
\label{sec:nmpc_position_control}
\subsection{System Dynamics}
\label{subsec:model}
Consider a body-fixed frame $\{\mathcal{B}\}$ located at the center of mass (CoM) of a quadrotor and aligned with a North-West-Up inertial frame $\{\mathcal{I}\}$. Then consider a quadrotor with position $p = (x, y, z)^T \in \mathbb{R}^3$ expressed in $\{\mathcal{I}\}$, attitude $ q = (q_w, q_x, q_y, q_z)^T \in \mathbb{H}$, linear velocity $v_b = (v_x, v_y, v_z)^T \in \mathbb{R}^3$ expressed in $\{\mathcal{B}\}$ and angular rate $\omega = (\omega_x, \omega_y, \omega_z)^T \in \mathbb{R}^3$. Its nonlinear dynamics are given by the first order ordinary differential equations adapted from \cite{luis2016}: 
\begin{equation}
    \dot{\xi} = f(\xi,u) = \begin{bmatrix}
    	Sv_b \\
    	\frac{1}{2}q \otimes \omega\\
    	\frac{1}{m}F_b - S^{T}g\mathbf{1_z} - \omega\times v_b\\
    	J^{-1}(M_b - \omega\times J\omega)
    \end{bmatrix} \label{eq:cont_dynamics},
\end{equation}
with state $\xi := (p, q, v_b,\omega)^T \in \mathbb{R}^{13}$. Here, $\mathbf{1_z} = (0, 0, 1)^T$, while the constants $g$ and $m$ represent the gravitational acceleration and the mass of the quadrotor, respectively. The inertia moment of the vehicle is denoted by $J = \text{diag}(J_{xx}, J_{yy}, J_{zz}) \in \mathbb{R}^{3 \times 3}$. The quaternion rotation matrix from  $\{\mathcal{B}\}$ to $\{\mathcal{I}\}$ is expressed by $S \vcentcolon \mathbb{H} \rightarrow \mathbb{R}^{3\times 3}$. The total external forces and moments applied to the CoM of the quadrotor and expressed in $\{\mathcal{B}\}$ are defined as: 
 \begin{equation*}
 	\begin{aligned}
    F_b & \vcentcolon =  \sum_{i=1}^4 C_T\Omega_{i}^2\mathbf{1_z}, \quad M_b \vcentcolon = (M_x, M_y, M_z)^T
	\end{aligned}
 \end{equation*}
 with
 \begin{equation*}
 	\begin{aligned}
    M_x & = C_T\cdot l(-\Omega_{1}^{2}-\Omega_{2}^{2}+\Omega_{3}^{2}+\Omega_{4}^{2})\\
    M_y & = C_T\cdot l(-\Omega_{1}^{2}+\Omega_{2}^{2}+\Omega_{3}^{2}-\Omega_{4}^{2})\\
    M_z & = C_D(-\Omega_{1}^{2}+\Omega_{2}^{2}-\Omega_{3}^{2}+\Omega_{4}^{2}),
 	\end{aligned}
 \end{equation*}
where $C_T$ is the thrust coefficient, $C_D$ is the drag coefficient, and $l$ is half of the distance between motors. Finally, assuming it is possible to change the rotational speed of the $i^{\text{th}}$ propeller $\Omega_i$ instantaneously, we define their set as the control input of the system:
\begin{equation*}
	u \vcentcolon = (\Omega_{1},\Omega_{2},\Omega_{3},\Omega_{4})^T \in \mathbb{R}^4.\label{eq:control_inputs}
\end{equation*}
\subsection{State Predictor}
\label{subsec:predictor}
In our aerial system, because of the radio communication latency, we have delays both in receiving measurements ($\tau_1$) and sending control inputs ($\tau_2$). Likewise, since we intend to use NMPC, the potentially high computational burden ($\tau_c$) associated with its solution becomes an element that must also be taken into account to minimize the error in the state prediction. 

One-way latency is inherently difficult to measure due to the synchronization clocks dilemma, an issue that is usually solved by measuring the \emph{round-trip time} (RTT). In this context, the sum of network latencies must be considered and can be deliberately placed where it is most convenient, as long as the RTT is not changed. Following these considerations, we propose a state predictor based on the RTT as a delay compensator. The prediction is computed by performing forward iterations of system \eqref{eq:cont_dynamics}, starting from the current measured state and over the RTT, through an explicit Runge Kutta 4th order (ERK4) integrator. Due to the independent nature of this operation, perfect delay compensation can be achieved by adjusting the integration step to be equal to the RTT. Thus, in this paper, it is assumed that there is a fixed RTT, defined by $\tau_r := \tau_1+\tau_2+\tau_c$, to be compensated. Although this prediction is simple (and inaccurate), it will prove surprisingly effective in experimentation.
\subsection{Nonlinear Model Predictive Control}
\label{subsec:nmpc}
Employing direct multiple shooting \cite{bock1984} to discretize the underlying continuous-time optimal control problem and assuming a linear least squares objective, the NMPC problem considered in this paper is defined as the following constrained nonlinear program (NLP):
\begin{equation}\label{eq:ocp}
{\!\!\!\!\!}{\!\!}\begin{aligned}
&\underset{\begin{subarray}{c}
\xi_0, \dots, \xi_N, \\
u_0, \dots, u_{N-1}
\end{subarray}}{\min}	    \frac{1}{2}\sum_{i=0}^{N-1} \resizebox{.29\hsize}{!}{$\|\eta(\xi_i, u_i)-\Tilde{\eta}_i\|^2_{W}$} \,\resizebox{.02\hsize}{!}{+}\, \frac{1}{2} \resizebox{.3\hsize}{!}{$\|\eta_N(\xi_N)-\Tilde{\eta}_N\|^2_{W_N}$}\\ 
&\,\,\,\quad \text{s.t.} \,\,\,\,\,\,\,\,\, \xi_0 = \Hat{\xi}(k+\tau_r), \\
&\,\,\,\,\,\,\,\,\,\,\,\,\,\,\,\,\,\,\,\,\,\,\,\,\, \xi_{i+1} - F(\xi_i,u_i) = 0, \,\,\,\,\,\, i = 0,\dots, N-1,\\
& \,\,\,\,\,\,\,\,\,\,\,\,\, \,\,\,\,\,\,\,\,\,\,\,\, u^l \leq u_i \leq u^u, \,\,\,\,\,\,\,\,\,\,\,\,\,\,\,\,\,\,\,\,\,\,\,\, i = 0,\dots, N-1,
\end{aligned}{\!\!\!}
\end{equation}
where $\xi := (\xi_0, \dots, \xi_N)^T$ and $u := (u_0, \dots, u_{N-1})^T$ denote the state and input trajectories of the discrete-time system whose dynamics are described by $F \vcentcolon \mathbb{R}^{n_{\xi}} \times \mathbb{R}^{n_u} \rightarrow \mathbb{R}^{n_{\xi}}$. The functions in the stage and terminal least squares terms are denoted by $\eta \vcentcolon \mathbb{R}^{n_{\xi}} \times \mathbb{R}^{n_u} \rightarrow \mathbb{R}$ and $\eta_N \vcentcolon \mathbb{R}^{n_{\xi}} \rightarrow \mathbb{R}$ respectively. Variables $\Tilde{\eta}_i \in \mathbb{R}^{(n_\xi + n_u)}$ and $\Tilde{\eta}_N \in \mathbb{R}^{n_\xi}$ denote the time-varying references. The weight matrices are denoted by $W,W_N \succ 0$. The inputs bounds are $u^l < u^u \in \mathbb{R}^{n_u}$. Finally, $N$ and $\Hat{\xi}(k+\tau_r)$ represent the horizon length and the estimated state of the system at the current time instant $k$, respectively. 

When using NMPC to control a system, at each sampling time, an instance of NLP \eqref{eq:ocp} needs to be solved within the available computation time, where the value of the current estimate of the system is set to $\Hat{\xi}(k+\tau_r)$. To this end, several numerical approaches have been investigated in the literature \cite{Diehl2009c}. Among others, two main classes of approaches stand out: sequential quadratic programming (SQP) and nonlinear interior-point (IP) methods.
\subsection{The Real-Time Iteration Scheme and \texttt{acados}}
\label{subsec:rti}
Due to the non-negligible computational burden associated with the solution of \eqref{eq:ocp}, it is often necessary to use approximate feedback policies to achieve real-time feasibility. A rather widespread approach, the so-called \emph{real-time iteration} (RTI) \cite{Diehl2005}, consists of using an SQP-based numerical strategy that relies on the solution of a limited number of QP subproblems. In this way, the computation times can be drastically reduced. Although only an approximate solution for \eqref{eq:ocp} is obtained, the parametric manifold that describes the optimal solution $u^{\star}(\xi)$ can be tracked, leading to a feedback policy that, under suitable assumptions \cite{Diehl2007b}, achieves bounded suboptimality and preserves attractivity of equilibria. In this work, we will be using the implementation of the RTI method through the high-performance software package \texttt{acados}.
\subsection{Condensing Approach and Structure-Exploiting QP Solver}
Among other methods to approximate the Hessian of the Lagrangian, a particularly successful one is the Generalized Gauss-Newton (GGN) method, which is applicable when the objective function is of least squares type \cite{bock1984}, as in  \eqref{eq:ocp}. After applying it, the resulting QP has a sparse banded structure \cite{kouzoupis2015} and can be rewritten as:
\begin{equation}\label{eq:qp}
{\!\!\!\!\!}{\!\!}\begin{aligned}
&\underset{\begin{subarray}{c}
w
\end{subarray}}{\min}	    &&\frac{1}{2}\sum_{i=0}^{N} w_i^TH_iw_i + h_i^Tw_i\\ 
&\,\text{s.t.}    &&\Delta\xi_0 = \Hat{\xi}(k+\tau_r) - \xi_0^n, \\
& 						    &&E_{i+1}w_{i+1} - C_iw_i - d_i = 0, \,\,\, i = 0,\dots, N-1,\\
& 						    &&g_i^l \leq G_iw_i \leq g_i^u, \,\,\,\,\,\,\,\,\,\,\,\,\,\,\,\,\,\,\,\,\,\,\,\,\,\,\,\,\,\,\, i = 0,\dots, N-1,
\end{aligned}{\!\!\!}
\end{equation}
where $w_i = [\Delta\xi_i^T \,\, \Delta u_i^T]^T, \,  w_N = \Delta \xi_N$ with state and control deviations $\Delta \xi_i = \xi_i-\xi_i^n, \, \Delta u_i = u_i - u_i^n$. To match the QP formulation used in \texttt{acados}, we define: $C_i \vcentcolon = [A_i \,\, B_i]$, $E_i \vcentcolon = [\mathbf{I} \,\,\, 0]$, $G_i \vcentcolon = [0 \,\,\, \mathbf{I}]$, for $i = 0, \dots, N-1$, and $E_N \vcentcolon = \mathbf{I}$. The matrices used in the linearized dynamics and the upper and lower bounds of the polyhedral constraints are defined as
\begin{equation*}
\begin{aligned}
	A_i &\vcentcolon = \nabla_\xi F(\xi_i^n,u_i^n)^T,\quad B_i \vcentcolon = \nabla_u F(\xi_i^n, u_i^n)^T, \\
	d_i &\vcentcolon = F(\xi_i^n,u_i^n) - A_i\xi_i^n - B_iu_i^n, \quad  \\
	g_i^l &\vcentcolon = u^l - \mathbf{I}u_i^n, \quad g_i^u \vcentcolon = u^u - \mathbf{I}u_i^n,
\end{aligned}
\end{equation*}
where the superscript $n$ refers to the linearization points at the previous QP iteration. Finally, we have that the Hessian and the gradient terms for $i = 0, \dots, N-1$ are defined as
\begin{equation*}
	H_i \vcentcolon = W_i = \begin{bmatrix}Q_i & 0\\0 & R_i\end{bmatrix}, \quad h_i \vcentcolon = \begin{bmatrix}q_i\\r_i\end{bmatrix}, 
\end{equation*}
where 
\begin{equation*}
	\begin{aligned}
		Q_i & = \nabla_{\xi}(\eta(\xi_i^n,u_i^n)-\Tilde{\eta}_i^n)W^\xi\nabla_{\xi}(\eta(\xi_i^n,u_i^n)-\Tilde{\eta}_i^n)^T \\
		R_i & = \nabla_{u}(\eta(\xi_i^n,u_i^n)-\Tilde{\eta}_i^n)W^u\nabla_{u}(\eta(\xi_i^n,u_i^n)-\Tilde{\eta}_i^n)^T \\
		q_i & = \nabla_{\xi}(\eta(\xi_i^n,u_i^n)-\Tilde{\eta}_i^n)(\eta(\xi_i^n,u_i^n)-\Tilde{\eta}_i^n) \\
		r_i & = \nabla_{u}(\eta(\xi_i^n,u_i^n)-\Tilde{\eta}_i^n)(\eta(\xi_i^n,u_i^n)-\Tilde{\eta}_i^n),
	\end{aligned}
\end{equation*}
with $W^\xi$ and $W^u$ denoting the blocks in $W$ related to states and inputs respectively, and 
\begin{equation*}
	\begin{aligned}
		Q_N & = \nabla_{\xi}(\eta_N(\xi_N^n)-\Tilde{\eta}_N^n)W_N\nabla_{\xi}(\eta_N(\xi_N^n)-\Tilde{\eta}_N^n)^T \\
		q_N & = \nabla_{\xi}(\eta_N(\xi_N^n)-\Tilde{\eta}_N^n)(\eta_N(\xi_N^n)-\Tilde{\eta}_N^n).
	\end{aligned}
\end{equation*}

One way to address the QP \eqref{eq:qp} could be to exploit its banded structure and apply a sparse QP solver to it. Another way could be to first reduce, or condense, the variable space of the QP by eliminating all state deviations through the continuity constraint of \eqref{eq:qp} and then apply a dense QP solver. Differently, in this paper, we employ a recently proposed approach, in-between the sparse and the condensed one, called \emph{partial condensing}. In particular, we use a Hessian condensing algorithm where a state component is retained as an optimization variable at each stage of the partially condensed QP \cite{frison2016}. The algorithm finds the optimal level of sparsity for the QP solver at hand, trading horizon length for input vector size. Thus, partial condensing is employed as a preparation step before the call to \texttt{HPIPM}, the Riccati-based interior-point method solver we use, which has been tailored to exploit the particular structure of the obtained QP.
%===============================================================================
\section{SIMULATION RESULTS}
\label{sec:simulations}
In this section, we will first assess the performance of two optimal controllers in simulation. In particular, a linear-quadratic regulator (LQR) and an NMPC will be considered. Such comparison is valuable since the LQR can perform an optimal control law within short execution times and with low computational burden, while the NMPC deals with constraints and predictions. The values of the parameters appearing in the model are listed in Table \ref{tab:physical_prm} and correspond to the parameters of the actual Crazyflie, obtained partly from the study in \cite{luis2016} and partly via system identification. Accounting for numerous computation time uncertainties, in this work, a sampling time of $\tau_s = 15$ ms is set for both controllers. Next, we will investigate the effect of the time-delay on the control performance of the nano-quadrotor system, considering different RTTs.

\begin{table}[h]
\caption{Physical parameters for the Crazyflie 2.1}
\label{tab:physical_prm}
\centering
\begin{tabular}{ccc}
\hline
Parameter & Value                                  & Description                               \\ \hline
$g$       & 9.8066 m/s$^2$                         & gravitational acceleration                \\ \hline
$m$       & 0.33 kg                                & total mass                                \\ \hline
$l$       & 0.0325 m                               & arm length    \\ \hline
$J_{xx}$    & $1.395\times 10^{-5}$ $\text{kg}{\cdot}\text{m}^2$ & inertia moment around x \\ \hline
$J_{yy}$    & $1.395\times 10^{-5}$ $\text{kg}{\cdot}\text{m}^2$ & inertia moment around y \\ \hline
$J_{zz}$    & $2.173\times 10^{-5}$ $\text{kg}{\cdot}\text{m}^2$ & inertia moment around z \\ \hline
$C_D$     & $7.9379 \times 10^{-6}$ Nm/krpm$^2$     & drag coefficient                          \\ \hline
$C_T$     & $3.25\times 10^{-4}$ N/krpm$^2$        & thrust coefficient                        \\ \hline
\end{tabular}
\end{table}
%===============================================================================
\subsection{LQR Controller}\label{subsec:lqr_sim}
The first controller considered is the LQR. During the LQR control design, the nonlinear dynamics \eqref{eq:cont_dynamics} are linearized around the hovering steady state and input ($\Bar{\xi},\Bar{u}$) and also discretized using $\tau_s$ as sampling time, producing:
\begin{equation}
	\xi_{k+1} - \Bar{\xi} = A(\xi_k - \Bar{\xi}) + B(u_k - \Bar{u}).\label{eq:local_model}%
\end{equation}
Due to the uncontrollable linearized quaternion dynamics, the local linear model  \eqref{eq:local_model} is also not controllable. To address this issue, we project the dynamics onto a controllable subspace as done in \cite{yang2012, zanelli2018}, using the fact that the first component of the quaternion vector can be eliminated thanks to the relation \scalebox{0.85}{$ q_w = \sqrt{1 - q_x^2 - q_y^2 - q_z^2}$}. Thus, we obtain the 12-state model needed to project the LQR static gain. A discrete-time LQR controller can be designed by finding the infinite horizon solution $P$ associated with the discrete-time algebraic Riccati equation
\begin{equation*}
	A^T PA-P-(A^T PB)(B^T PB+R)^{-1}(B^T PA) + Q = 0,
\end{equation*}
which yields the static feedback law $u = K\xi + \Bar{u}$, where
\begin{equation*}
	K = (B^T PB+R)^{-1}(B^T PA).\label{eq:lqr_gain}
\end{equation*}
The positive-definite matrices $Q$ and $R$ are defined so that the stability of the closed-loop system is guaranteed.
%===============================================================================
\subsection{NMPC Controller}\label{subsec:nmpc_sim}
The second controller is the NMPC. When designing an NMPC, choosing the horizon length has profound implications for computational burden and tracking performance. For the former, the longer the horizon, the higher the computational burden. As for the latter, in principle, a long prediction horizon tends to improve the overall performance of the controller. In order to select this parameter and achieve a trade-off between performance and computational burden, NLP \eqref{eq:ocp} has been implemented in \texttt{acados} considering $N =\{10,20,30,40,50\}$, discretizing the dynamics \eqref{eq:cont_dynamics} using an ERK4 integration scheme. Likewise, we compare the condensing approach with the state-of-the-art solver \texttt{qpOASES} \cite{ferrau2014} against the partial condensing approach with \texttt{HPIPM}, in relation to the set of horizons considered. The weight matrices $W, W_N$ are the same in all simulations. The functions in the stage and terminal least squares terms are defined as
\begin{equation}
\begin{aligned}[c]
\eta(\xi_i,u_i)  \vcentcolon = \begin{bmatrix}\xi_{i}\\u_{i}\end{bmatrix}
\end{aligned}
,\quad
\begin{aligned}[c]
\eta_N(\xi_N) \vcentcolon = \xi_{N} \label{residuals}.
\end{aligned}
\end{equation}
\begin{remark}
	We regulate for $q^d = (1,0,0,0)^T$ so that the geodesic arc can be locally approximated by a straight line.
\end{remark}
\begin{figure}[t]
    \centering
    \includegraphics[width=.5\textwidth]{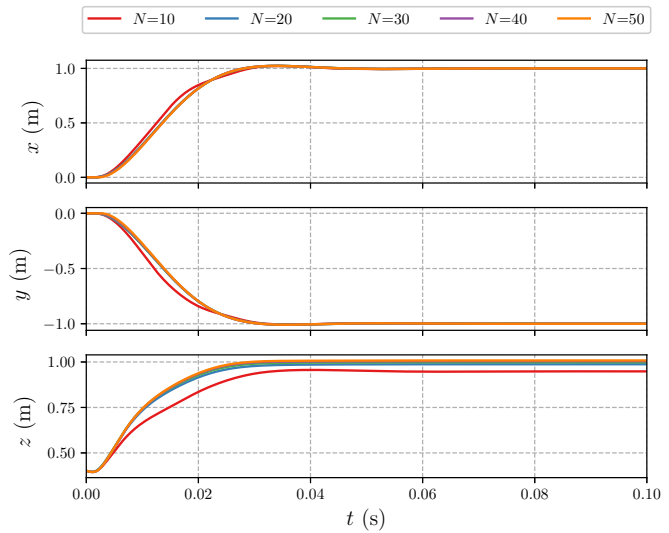}
    \vspace*{-7mm}
    \caption{Closed-loop trajectories comparing different horizon lengths.}\label{fig:horizon}%
\end{figure}

\begin{figure}[t]
    \centering
    \includegraphics[width=.4\textwidth]{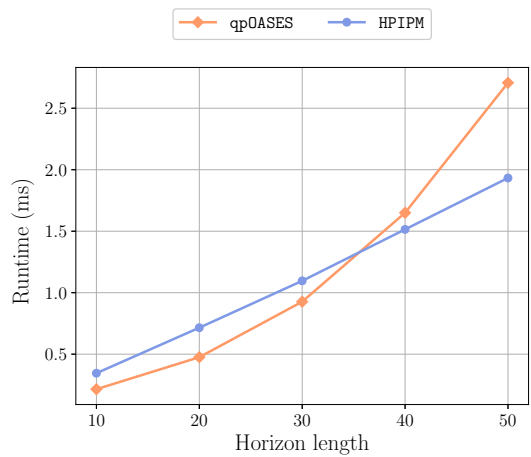} 
    \vspace*{-1.5mm}
    \caption{Average runtimes per SQP-iteration for five different horizons considering two distinct QP solvers for QP \eqref{eq:qp}.}\label{fig:cpu_info}%
\end{figure}

According to Fig. \ref{fig:horizon}, horizon $N=50$ provides a higher tracking performance. Nevertheless, when considering a long horizon, the available computational power may not be sufficient to deliver a solution within the runtime requirements of an aerial system. In Fig. \ref{fig:cpu_info}, the average runtimes per SQP-iteration are reported. As \texttt{qpOASES} is a solver based on active-set method, it requires condensing to be computationally efficient. In line with the observations found in the literature that condensing is effective for short to medium horizon lengths, we note that \texttt{qpOASES} is competitive for horizons up to approximately $N=30$ when compared to \texttt{HPIPM}. The break-even point moves higher on the scale for longer horizons, mainly due to efficient software implementations that cover: a) Hessian condensing procedure tailored for partial condensing, b) structure-exploiting QP solver based on novel Riccati recursion, c) hardware-tailored linear algebra library. 

Furthermore, when dealing with a nominal NMPC formulation, as in this paper, the model uncertainties and their propagation also contribute to the decision of the horizon length. The longer the prediction, the more the modeling errors accumulate, and the tracking performance can no longer be guaranteed. Uncertainty can be added to the system model to account for the noise associated with the onboard inertial measurement unit (IMU) and to simulate the variability in the repetition of the task. However, this has not been included in the scope of this paper. 

Therefore, we can conclude that horizon $N=50$ offers a reasonable trade-off between deviation from the reference trajectory and computational burden. 
%===============================================================================
\subsection{Comparison}\label{subsec:comparison}
\begin{figure}[t]
	\centering
	\includegraphics[width=.5\textwidth]{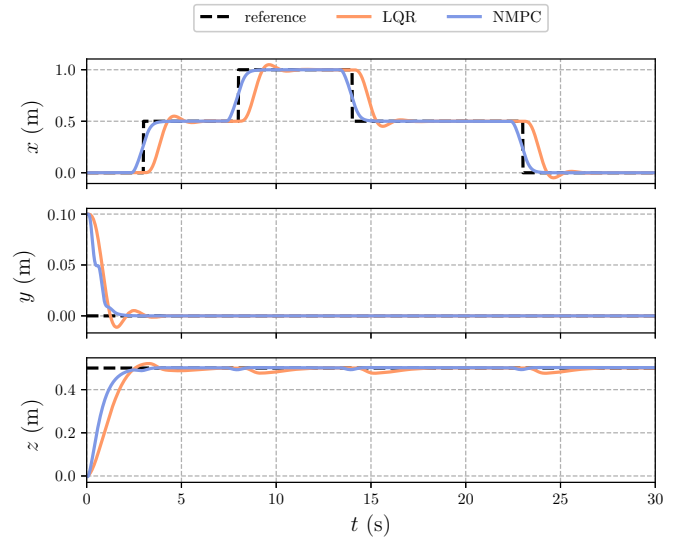}
	\vspace*{-7mm}
    \caption{Simulation results comparing different controllers: closed-loop trajectories for position tracking.}\label{fig:comparison}%
\end{figure}

This section is dedicated to compare and discuss the simulation results for LQR and NMPC. Initially, the LQR was tuned so that its control policy was locally equivalent to the one of the NMPC. However, as also experienced in the work of \cite{zanelli2018}, the simulations showed the need to detune the controller so that the performance, specifically for the position in $z$ in our case, was acceptable. Thus, the following weight matrices are used:
\begin{align*}
	Q &= \text{blkdiag}(12.24 \cdot 10^{3}, 10.2 \cdot 10^{3}, 9 \cdot 10^{5},0.102 \cdot \mathbf{I_3}, 71.4,\\
		&\quad \quad 102, 408, 1.02 \cdot 10^{-3}\cdot \mathbf{I_2}, 1.02\cdot 10^3),\, R = 0.12\cdot \mathbf{I_4}.
\end{align*}
For the NMPC, the weight matrices are
\begin{align*}
    W &= \text{blkdiag}(120,100\cdot \mathbf{I_2},1\cdot 10^{-3} \cdot \mathbf{I_4}, 7 \cdot 10^{-1},1,4,\\
    & \quad \quad \quad \quad\,\,\,\, 1\cdot 10^{-5}\cdot \mathbf{I_2},10,6\cdot 10^{-2} \cdot \mathbf{I_4}),\\
    W_N &= 50\cdot \text{blkdiag}(120,100\cdot \mathbf{I_2},1\cdot 10^{-3} \cdot \mathbf{I_4}, 7 \cdot 10^{-1},1,4,\\
    & \quad \quad \quad \quad \quad \quad \,\,\, 1\cdot 10^{-5}\cdot \mathbf{I_2},10),%
\end{align*}
while the input bounds on the rotational speed of the propellers are $u^l = 0$, $u^u = 22$ krpm. The simulation scenario consists of producing steep changes in the position reference. The closed-loop tracking performance of both controllers is shown in Fig.~\ref{fig:comparison}. We notice that the NMPC controller outperforms the LQR, considering that absence of overshoots and faster response to changes in the references can be achieved. This may be due to the fact that nonlinearity and unmodeled constraints can degrade the performance of the LQR controller for large reference changes. 

%===============================================================================
\subsection{Time-Delay}
\begin{figure}[t]
	\centering
	\includegraphics[width=.45\textwidth]{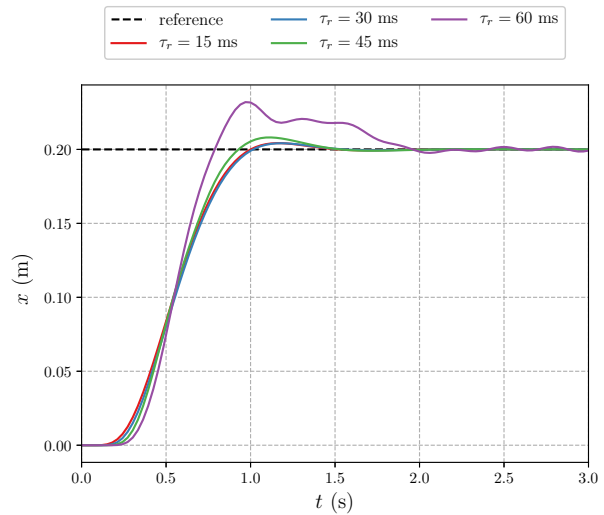}
	\vspace*{-3mm}
    \caption{Crazyflie NMPC step response for four distinct RTTs without considering time-delay compensation.}\label{fig:delay}%
\end{figure}

For the purposes of this analysis, we assume that $\tau_r = \lambda\tau_s$ for some nonnegative integer $\lambda$. However, we emphasize that the same analysis can be done considering any value of $\tau_r$, as mentioned in Section \ref{subsec:predictor}. The results of the simulations for the step response of proposed NMPC without considering the time-delay compensation are shown in Fig.~\ref{fig:delay}. We observe that the greater the time-delay associated with the RTT, the greater the degradation on the NMPC performance. This degradation is observed through the increase of overshoots and oscillations in the system's response, indicating that the proposed NMPC is not able to control the system properly. Unlike in simulation, in a real-world context, where model uncertainties come into view, the overshoots and oscillations will tend to be even greater. Besides, other simulations have shown that the system becomes unstable for greater RTTs. That is why delay compensation is of particular importance in our case.

%===============================================================================
%===============================================================================
\section{EXPERIMENTAL RESULTS}
\label{sec:experiments}
\begin{figure*}[t]
\centering
 \includegraphics[width=1.0\textwidth,height=4cm]{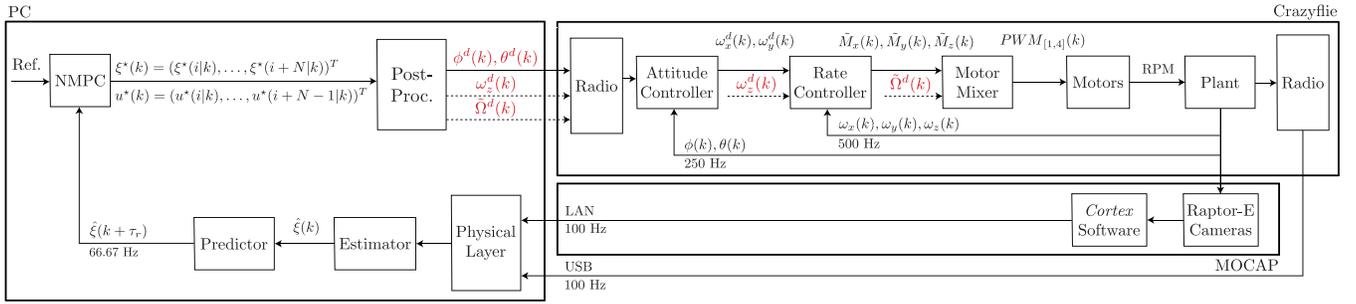}
 \caption{Schematics of the onboard controllers, motion capture (MOCAP) system, and the proposed offboard control architecture.}
 \label{fig:block_diagram}
\end{figure*}
%===============================================================================
\subsection{Setup}
The proposed architecture is experimentally validated on the Crazyflie 2.1 nano-quadrotor, developed by the Swedish company Bitcraze\footnote{\url{https://www.bitcraze.io}}. The system can be controlled through the onboard long-range radio receiver and its correspondent transmitter, the Crazyradio PA, or even through the Bluetooth LE (low-energy) connection. It has an open-source firmware based on FreeRTOS that allows the user to log sensor data at a maximum frequency of 100 Hz. The radio communication uses a protocol developed by Bitcraze, the Crazy RealTime Protocol (CRTP), which limits the size of the packages sent from the Crazyflie to the Crazyradio in up to 30 bytes (1-byte header, 29-bytes data payload).

The motion capture system consists of ten Raptor-E cameras and the \textit{Cortex} software, both from Motion Analysis. It provides the 3D global position of an IR-reflective marker placed on top of the Crazyflie at a frequency of 100 Hz. The entire setup has an estimated RTT of $60$ ms to be compensated.
%===============================================================================
\subsection{Software Interface}
Our control architecture hinges upon a \texttt{ROS Kinetic} framework and runs at 66.67 Hz. The CRTP is used in combination with our \texttt{crazyflie\_nmpc} stack \cite{github2020}, based on the Crazyflie ROS interface \cite{honig2017}, to stream in runtime custom packages containing the required data to reconstruct the part of the measurement vector that depends on the IMU data. The logging subsystem is mapped into ROS messages and, thus, it is possible to recover the current Euler angles and angular velocities of the nano-quadrotor. Likewise, the \texttt{cortex\_ros} bridge streams the 3D global position of the Crazyflie, which is then passed through a second-order, discrete-time Butterworth filter to estimate the linear velocities. 

Through the use of \texttt{acados} Python interface and \texttt{CasADi} automatic differentiation and modeling language, we export a plain C-code that corresponds to NLP \eqref{eq:ocp}. The exported code is linked against the precompiled \texttt{acados} core library and then wrapped by our NMPC ROS node, written in C++. The QP subproblems arising in the SQP algorithm in \texttt{acados} are solved using the high-performance \texttt{HPIPM} solver, which implements an interior-point method. It is written entirely in C and is built on top of the high-performance linear algebra package \texttt{BLASFEO}. We use the \texttt{X64\_INTEL\_HASWELL} target in \texttt{BLASFEO}, which explicitly employs the powerful vector instructions provided by this computer architecture. In particular, the high-performance routines provided by \texttt{BLASFEO} outperform other state-of-the-art dense linear algebra libraries in the case of matrices of moderate size, as in the current application. Regarding the condensing approach, we use the partial condensing routines provided by \texttt{HPIPM}.
%===============================================================================
\subsection{Implementation Considerations}
Although system \eqref{eq:cont_dynamics} is used in our NMPC formulation, the standard \textit{input commands} for the actual Crazyflie (indicated by the red variables in Fig. \ref{fig:block_diagram}) are:
\begin{itemize}
	\item the desired roll $\phi^d$ and pitch $\theta^d$ angles, in degrees
	\item the yaw rate $\omega_{z}^d$, in degrees per second
	\item the base PWM signal $\Tilde{\Omega}^d$
\end{itemize}
It is important to clarify that $\Tilde{\Omega}^d$ is a 16-bit integer number ranging from 0 to 65535 and stands for the base value of pulse width modulation (PWM) applied to all motors to maintain altitude. Despite the name \textit{thrust} in the official firmware\footnote{\url{https://github.com/bitcraze/crazyflie-firmware}}, $\Tilde{\Omega}^d$ does not directly represent the net force applied along the $z$ body-fixed axis, but is related to it. In principle, once the NMPC solution is obtained, the first element of the predicted control trajectory $u^\star(0|k)$ should be sent to the Crazyflie. However, there is an incompatibility between  $\Tilde{\Omega}^d$ and $u^\star(0|k)$. To translate the control inputs $\Omega_i^\star = u^\star(0|k)$, in krpm, to the base PWM signal $\Tilde{\Omega}^d$, an integer, we need to establish some mappings. To that end, we use the following relations adapted from \cite{luis2016}:
\begin{equation*}
\begin{aligned}[c]
	\Omega^d  = \frac{1}{4}\sum_{i=1}^4 \Omega_i^\star \text{ [krpm]}
\end{aligned}
,\quad
\begin{aligned}[c]
	\Tilde{\Omega}^d  = \frac{1000\cdot \Omega^d-4070.3}{0.2685} \text{ [int]}.
\end{aligned}
\end{equation*}
Similarly, at each new NMPC solution, we reconstruct the remaining input commands $\phi^d$, $\theta^d$, $\omega_{z}^d$ using part of the state solution $\xi^{\star}(i|k)$ at stage $i=1$, as denoted in Fig. \ref{fig:block_diagram}. This approach, however, entails some distortions onto the onboard controllers (attitude and rate controllers) since the NMPC does not take them into account in its formulation. This effect will be described in the next subsection.
%===============================================================================
\subsection{Onboard Controller Considerations}
How the onboard controllers (PIDs) use the setpoints of the offboard controller (NMPC) in our architecture is not entirely conventional and, thereby, deserves some considerations. Two main points presented in Fig. \ref{fig:block_diagram} must be addressed. First, the reference signals that the PID loops track do not fully correspond to the control inputs considered in the NMPC formulation. Instead, part of the state solution $\xi^\star(1|k)$ is used in conjunction with the control inputs $u^\star(0|k)$ to reconstruct the actual input commands passed as a setpoint to the Crazyflie. Second, a part of the reconstructed input commands is sent as a setpoint to the outer loop (attitude controller), and the other part is sent to the inner loop (rate controller).  

Furthermore, as the NMPC model does not include the PID loops, it does not truly represent the real system, even in the case of perfect knowledge of the physical parameters. As a consequence, the optimal feedback policy is distorted in the real system by the PIDs. Note that intuitively reducing the NMPC discretization time mitigates the distortion. Another solution is to implement a linear interpolation onboard, considering the frequency of each loop. In this case, given a new input command from the NMPC, data points are constructed and passed as reference signals to the PIDs.
%===============================================================================
\subsection{Experiments Description}
\newcolumntype{b}{X}
\newcolumntype{s}{>{\hsize=.4\hsize}X}

\begin{figure*}[t]
\begin{tabularx}{\textwidth}{X}
\centering
\subfloat[Smooth step.]{\label{fig:smooth_step}\includegraphics[width=.9\textwidth]{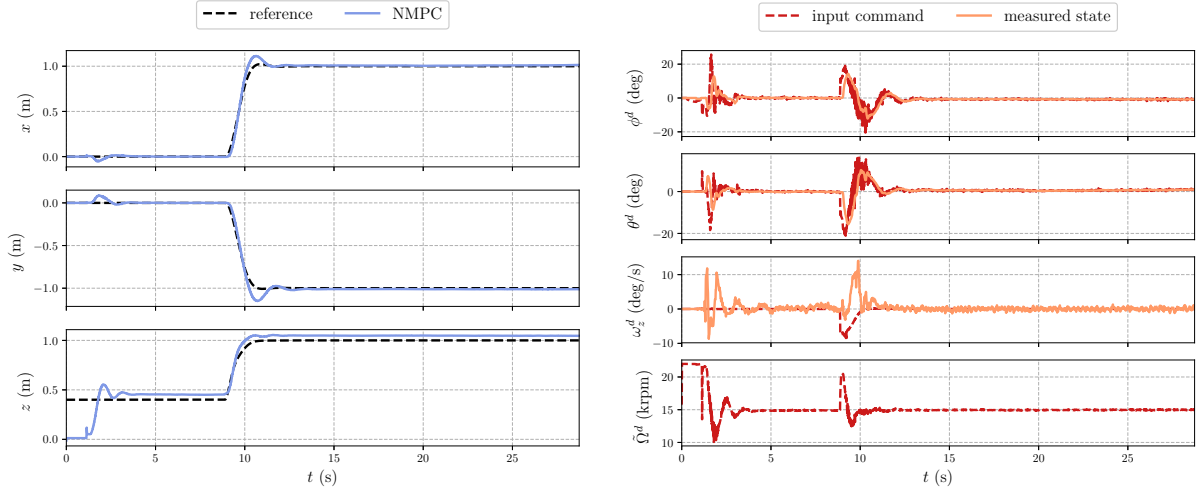}}
\\%
\subfloat[Helical reference.]{\label{fig:helical}\includegraphics[width=.9\textwidth]{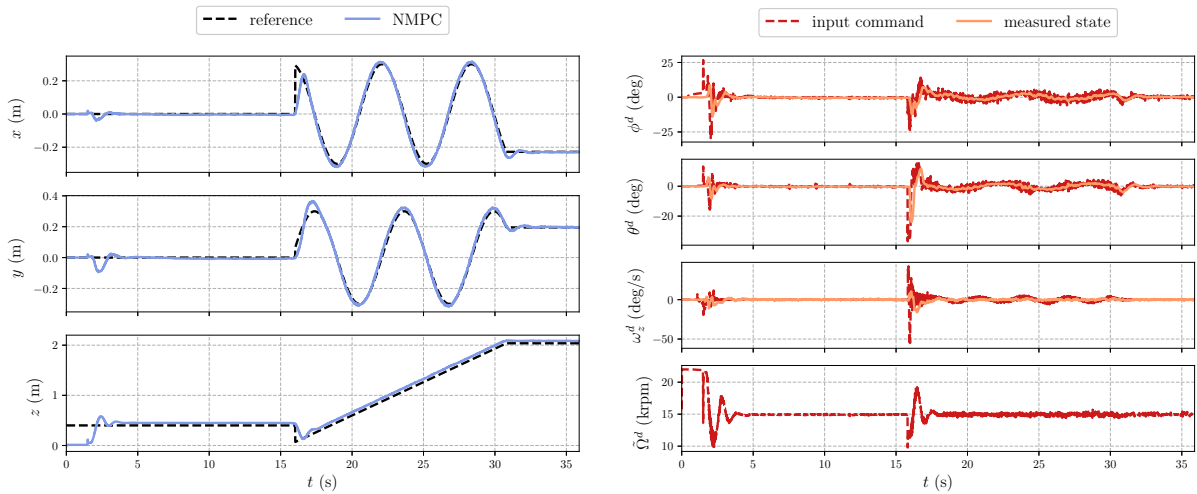}}
\end{tabularx}
\caption{Experimental results: output trajectory and input commands.}\label{fig:closed_loop}
\end{figure*}

In every experiment, the Crazyflie takes off and remains hovering at $p^{d} = (0,0,0.4)^T$. This is done to avoid the unmodeled ground effect during the tracking phase. When the nano-quadrotor reaches the steady-state for hovering, the user triggers the tracking. The nano-quadrotor is then asked to hold the last point of the reference trajectory until the user triggers the landing command.

To validate the effectiveness of our control architecture, we ran two experiments. For each experiment, we generate a reference trajectory on a base computer and pass it to our NMPC ROS node every $\tau_s = 15$ ms. When generating the trajectories, we explicitly address the feasibility issue in the design process, creating two references: one feasible and one infeasible. In addressing this issue, we prove through experiments that the performance of the proposed NMPC is not degraded even when the nano-quadrotor attempts to track an infeasible trajectory, which could, in principle, make it deviate significantly or even crash. The reference trajectories considered for each experiment are:

\subsubsection{Smooth Step}
For the first experiment, the reference is a dynamically feasible smooth step on $(x, y, z)$ that drives the nano-quadrotor from the hover position to the final one, defined as $p^{d} \vcentcolon = (1,{-1},1)^T$. To generate it, we implemented NLP \eqref{eq:ocp} using \texttt{CasADi} with a time horizon of $T = 6$ s and $N = 400$ shooting nodes, discretizing the dynamics \eqref{eq:cont_dynamics} using an ERK4 integration method. The resulting NLP is solved using the interior-point solver \texttt{IPOPT} \cite{wachter2006}. The closed-loop trajectory and input commands for this experiment are shown in Fig. \ref{fig:smooth_step}. 

\subsubsection{Helical Reference} 
For the second experiment, the reference is a cartesian helical trajectory, generated using MATLAB. We imposed a radius of $r = 0.3$ m, initial height of $h_0 = 0.38$ m, height increments of $\Delta h = 0.002$ m per $\tau_s$, and a duration of $t_f = 15$ s, divided into $m = 1000$ intervals. The closed-loop trajectory and input commands for this experiment are presented in Fig. \ref{fig:helical}. 
%===============================================================================
%===============================================================================
\section{DISCUSSION}
\label{sec:discussion}
In Fig. \ref{fig:closed_loop}, we note that the PID distortion on the NMPC performance is relatively small. As expected, the most challenging setpoints to be tracked are the positions in which, given a change in the motion, the Crazyflie has to pitch/roll in the opposite direction quickly. These are the setpoints where the distortion has the greatest influence on the system, causing the small overshoots in position.

To assess the computational complexity of our algorithm, we gather the average and maximum solution times of the tailored RTI scheme using \texttt{acados}. The results were obtained on an Intel Core i5-8250U @3.4 GHz running Ubuntu. We had $t_{\text{AVG}} = 7.4$ ms and $t_{\text{MAX}} = 22.1$ ms where the computational peak was during take-off. Throughout the tracking phase, the computational workload offered a reasonable safety margin within the sampling time. These results show the efficiency of the proposed scheme.
%===============================================================================
\section{CONCLUSIONS}
\label{sec:conclusion}
In this paper, we presented the design and implementation of a novel position controller based on nonlinear model predictive control for quadrotors. The control architecture incorporates a predictor as a delay compensator for granting a delay-free model in the NMPC formulation, which in turn enforces bounds on the actuators. To validate our architecture, we implemented it on the Crazyflie 2.1 nano-quadrotor. The experiments demonstrate that the efficient RTI-based scheme, exploiting the full nonlinear model, achieves a high-accuracy tracking performance and is fast enough for real-time deployment. 

%\addtolength{\textheight}{-12cm}
%===============================================================================

\bibliographystyle{IEEEtran}
\bibliography{biblio}

\begin{thebibliography}{10}
\providecommand{\url}[1]{#1}
\csname url@samestyle\endcsname
\providecommand{\newblock}{\relax}
\providecommand{\bibinfo}[2]{#2}
\providecommand{\BIBentrySTDinterwordspacing}{\spaceskip=0pt\relax}
\providecommand{\BIBentryALTinterwordstretchfactor}{4}
\providecommand{\BIBentryALTinterwordspacing}{\spaceskip=\fontdimen2\font plus
\BIBentryALTinterwordstretchfactor\fontdimen3\font minus
  \fontdimen4\font\relax}
\providecommand{\BIBforeignlanguage}[2]{{%
\expandafter\ifx\csname l@#1\endcsname\relax
\typeout{** WARNING: IEEEtran.bst: No hyphenation pattern has been}%
\typeout{** loaded for the language `#1'. Using the pattern for}%
\typeout{** the default language instead.}%
\else
\language=\csname l@#1\endcsname
\fi
#2}}
\providecommand{\BIBdecl}{\relax}
\BIBdecl

\bibitem{deng2020}
H.~{Deng} and T.~{Ohtsuka}, ``{ParNMPC} -- a parallel optimization toolkit for
  real-time nonlinear model predictive control,'' \emph{International Journal
  of Control}, pp. 1--32, 2020.

\bibitem{barroscarlos2020}
B.~Barros~Carlos, T.~Sartor, A.~Zanelli, M.~Diehl, and G.~Oriolo, ``Least
  conservative linearized constraint formulation for real-time motion
  generation,'' in \emph{Proc. 21st IFAC World Congress}, 2020, pp. 9519--9525.

\bibitem{englert2020}
T.~Englert and K.~Graichen, ``Nonlinear model predictive torque control and
  setpoint computation of induction machines for high performance
  applications,'' \emph{Control Engineering Practice}, vol.~99, p. 104415,
  2020.

\bibitem{sathya2020}
A.~S. {Sathya}, J.~{Gillis}, G.~{Pipeleers}, and J.~{Swevers}, ``Real-time
  robot arm motion planning and control with nonlinear model predictive control
  using augmented lagrangian on a first-order solver,'' in \emph{Proc. European
  Control Conference (ECC)}, 2020, pp. 507--512.

\bibitem{guo2020}
N.~Guo, B.~Lenzo, X.~Zhang, Y.~Zou, R.~Zhai, and T.~Zhang, ``A real-time
  nonlinear model predictive controller for yaw motion optimization of
  distributed drive electric vehicles,'' \emph{IEEE Transactions on Vehicular
  Technology}, vol.~69, no.~5, pp. 4935--4946, 2020.

\bibitem{katayama2020}
S.~Katayama, M.~Doi, and T.~Ohtsuka, ``A moving switching sequence approach for
  nonlinear model predictive control of switched systems with state-dependent
  switches and state jumps,'' \emph{International Journal of Robust and
  Nonlinear Control}, vol.~30, no.~2, pp. 719--740, 2020.

\bibitem{kartal2020}
Y.~Kartal, K.~Subbarao, N.~R. Gans, A.~Dogan, and F.~Lewis, ``Distributed
  backstepping based control of multiple {UAV} formation flight subject to time
  delays,'' \emph{IET Control Theory \& Applications}, vol.~14, no.~12, pp.
  1628--1638, 2020.

\bibitem{alvarez2019}
J.~Alvarez-Mu{\~n}oz, J.~Castillo-Zamora, J.~Escareno, I.~Boussaada,
  F.~M{\'e}ndez-Barrios, and O.~Labbani-Igbida, ``Time-delay control of a
  multi-rotor {VTOL} multi-agent system towards transport operations,'' in
  \emph{Proc. International Conference on Unmanned Aircraft Systems (ICUAS)},
  2019, pp. 276--283.

\bibitem{huang2019}
Z.~Huang, Y.-J. Pan, and R.~Bauer, ``Leader--follower consensus control of
  multiple quadcopters under communication delays,'' \emph{Journal of Dynamic
  Systems, Measurement, and Control}, vol. 141, no.~10, 2019.

\bibitem{verschueren2019}
\BIBentryALTinterwordspacing
R.~Verschueren, G.~Frison, D.~Kouzoupis, N.~van Duijkeren, A.~Zanelli,
  B.~Novoselnik, J.~Frey, T.~Albin, R.~Quirynen, and M.~Diehl, ``acados: a
  modular open-source framework for fast embedded optimal control,'' 2019.
  [Online]. Available: \url{https://arxiv.org/abs/1910.13753}
\BIBentrySTDinterwordspacing

\bibitem{andersson2019}
J.~A. Andersson, J.~Gillis, G.~Horn, J.~B. Rawlings, and M.~Diehl, ``{CasADi}:
  a software framework for nonlinear optimization and optimal control,''
  \emph{Mathematical Programming Computation}, vol.~11, no.~1, pp. 1--36, 2019.

\bibitem{frison2020}
\BIBentryALTinterwordspacing
G.~Frison and M.~Diehl, ``{HPIPM}: a high-performance quadratic programming
  framework for model predictive control,'' 2020. [Online]. Available:
  \url{https://arxiv.org/abs/2003.02547}
\BIBentrySTDinterwordspacing

\bibitem{frison2018}
G.~Frison, D.~Kouzoupis, T.~Sartor, A.~Zanelli, and M.~Diehl, ``{BLASFEO}:
  Basic linear algebra subroutines for embedded optimization,'' \emph{ACM
  Transactions on Mathematical Software (TOMS)}, vol.~44, no.~4, p.~42, 2018.

\bibitem{luis2016}
\BIBentryALTinterwordspacing
C.~Luis and J.~L. Ny, ``Design of a trajectory tracking controller for a
  nanoquadcopter,'' 2016. [Online]. Available:
  \url{https://arxiv.org/abs/1608.05786}
\BIBentrySTDinterwordspacing

\bibitem{bock1984}
H.~G. Bock and K.~J. Plitt, ``A multiple shooting algorithm for direct solution
  of optimal control problems,'' \emph{9th IFAC World Congress}, vol.~17,
  no.~2, pp. 1603--1608, 1984.

\bibitem{Diehl2009c}
M.~Diehl, H.~J. Ferreau, and N.~Haverbeke, ``Efficient numerical methods for
  nonlinear {MPC} and moving horizon estimation,'' in \emph{Nonlinear model
  predictive control}, ser. Lecture Notes in Control and Information Sciences,
  L.~Magni, M.~Raimondo, and F.~Allg\"ower, Eds.\hskip 1em plus 0.5em minus
  0.4em\relax Springer, 2009, vol. 384, pp. 391--417.

\bibitem{Diehl2005}
M.~Diehl, H.~G. Bock, and J.~P. Schl\"oder, ``A real-time iteration scheme for
  nonlinear optimization in optimal feedback control,'' \emph{{SIAM} Journal on
  Control and Optimization}, vol.~43, no.~5, pp. 1714--1736, 2005.

\bibitem{Diehl2007b}
M.~Diehl, R.~Findeisen, and F.~Allg\"ower, ``A stabilizing real-time
  implementation of nonlinear model predictive control,'' in \emph{Real-Time
  and Online PDE-Constrained Optimization}, L.~Biegler, O.~Ghattas,
  M.~Heinkenschloss, D.~Keyes, and B.~van Bloemen~Waanders, Eds.\hskip 1em plus
  0.5em minus 0.4em\relax {SIAM}, 2007, pp. 23--52.

\bibitem{kouzoupis2015}
D.~Kouzoupis, R.~Quirynen, J.~Frasch, and M.~Diehl, ``Block condensing for fast
  nonlinear {MPC} with the dual {Newton} strategy,'' \emph{IFAC-PapersOnLine},
  vol.~48, no.~23, pp. 26--31, 2015.

\bibitem{frison2016}
G.~Frison, D.~Kouzoupis, J.~B. J{\o}rgensen, and M.~Diehl, ``An efficient
  implementation of partial condensing for nonlinear model predictive
  control,'' in \emph{Proc. Conference on Decision and Control (CDC)}, 2016,
  pp. 4457--4462.

\bibitem{yang2012}
Y.~Yang, ``Spacecraft attitude determination and control: Quaternion based
  method,'' \emph{Annual Reviews in Control}, vol.~36, no.~2, pp. 198--219,
  2012.

\bibitem{zanelli2018}
A.~Zanelli, G.~Horn, G.~Frison, and M.~Diehl, ``Nonlinear model predictive
  control of a human-sized quadrotor,'' in \emph{Proc. European Control
  Conference (ECC)}, 2018, pp. 1542--1547.

\bibitem{ferrau2014}
H.~Ferreau, C.~Kirches, A.~Potschka, H.~Bock, and M.~Diehl, ``{qpOASES}: A
  parametric active-set algorithm for quadratic programming,''
  \emph{Mathematical Programming Computation}, vol.~6, no.~4, pp. 327--363,
  2014.

\bibitem{github2020}
crazyflie\_nmpc, \url{https://github.com/bcbarbara/crazyflie_nmpc}, 2020.

\bibitem{honig2017}
W.~H{\"o}nig and N.~Ayanian, ``Flying multiple {UAVs} using {ROS},'' in
  \emph{Robot Operating System (ROS)}.\hskip 1em plus 0.5em minus 0.4em\relax
  Springer, 2017, pp. 83--118.

\bibitem{wachter2006}
A.~W{\"a}chter and L.~T. Biegler, ``On the implementation of an interior-point
  filter line-search algorithm for large-scale nonlinear programming,''
  \emph{Mathematical programming}, vol. 106, no.~1, pp. 25--57, 2006.

\end{thebibliography}

\end{document}